\documentclass{article}

\usepackage{PRIMEarxiv}

\usepackage[utf8]{inputenc} 
\usepackage[T1]{fontenc}    
\usepackage{hyperref}       
\usepackage{url}            
\usepackage{booktabs}       
\usepackage{amsfonts}       
\usepackage{nicefrac}       
\usepackage{microtype}      
\usepackage{lipsum}
\usepackage{fancyhdr}       
\usepackage{graphicx}       
\usepackage{natbib}
\usepackage{xcolor}
\usepackage{tcolorbox}
\usepackage{float}
\graphicspath{{media/}}     
\definecolor{darkgreen}{rgb}{0.0, 0.26, 0.15}

\title{LegiLM: A Fine-Tuned Legal Language Model for Data Compliance 
}

\author{
  Linkai Zhu, Lu Yang, Chaofan Li \\
  School of Management Science and Information Engineering\\
  Hebei University of Economics and Business\\
  Shijiazhuang,050061,China  \\
  \texttt{\{linkaizhu, yanglu,lichaofan\}hueb@edu.cn} \\
   \And
  Shanwen Hu \\
  Data Science Faculty  \\
  City University of Macau \\
  Macao, China\\
  \texttt{D23092100358@cityu.mo} \\
   \AND
   Lu Liu, Bin Yin \\
   QI-ANXIN Technology Research Institute  \\
   Beijing, China \\
   \texttt{\{liulu01, yinbin\}@qianxin.com}\\
}

\begin{document}
\maketitle

\begin{abstract}
Ensuring compliance with international data protection standards for privacy and data security is a crucial but complex task, often requiring substantial legal expertise. This paper introduces LegiLM, a novel legal language model specifically tailored for consulting on data or information compliance. LegiLM leverages a pre-trained GDPR Fines dataset and has been fine-tuned to automatically assess whether particular actions or events breach data security and privacy regulations. By incorporating a specialized dataset that includes global data protection laws, meticulously annotated policy documents, and relevant privacy policies, LegiLM is optimized for addressing data compliance challenges. The model integrates advanced legal reasoning methods and information retrieval enhancements to enhance accuracy and reliability in practical legal consulting scenarios. Our evaluation using a custom benchmark dataset demonstrates that LegiLM excels in detecting data regulation breaches, offering sound legal justifications, and recommending necessary compliance modifications, setting a new benchmark for AI-driven legal compliance solutions. Our resources are publicly available at https://github.com/DAOLegalAI/LegiLM
\end{abstract}


\section{Introduction}
Recent progress in large language models (LLMs) has led to notable advancements in model architecture and training techniques, showcasing the adaptability of LLMs across multiple fields. The GPT-4 report highlights a highly proficient model for producing human-like text, setting a new standard in text generation capabilities \cite{OpenAI2023}. Similarly, the GLM model employs autoregressive blank infilling to deliver strong performance across various tasks \cite{Du2022}, while LLaMA\cite{Touvron2023} focuses on efficiency and accessibility, especially under limited resources.

Despite these improvements, challenges such as interpretability, ethics, and bias continue to be significant concerns within the NLP community. Moreover, the introduction of comprehensive data protection laws has greatly increased the complexity of legal compliance, particularly in terms of privacy and data security. Legal professionals often face the meticulous and time-consuming task of reviewing relevant agreements to ensure they meet these stringent requirements.

Addressing this challenge, this paper introduces LegiLM, a meticulously fine-tuned legal language model specifically designed to automate the detection of violations in the realm of privacy and data security. This initiative represents a response to broader trends in NLP, where the potential of LLMs to revolutionize sectors like healthcare \cite{Singhal2023} and software development \cite{Tang2021, Muennighoff2023} has already been recognized. However, the legal field presents unique challenges that necessitate specialized applications \cite{Martin2024, Licari2022}. Initial forays into legal-focused LLMs have primarily concentrated on enhancing legal advice and document retrieval functionalities, but these early implementations often rely on the general capabilities of LLMs and do not sufficiently address the complex needs of legal reasoning and specific compliance tasks \cite{Yue2023, Chen2013}.

By leveraging advanced LLM technologies, LegiLM undergoes extensive pretraining on specialized legal corpora drawn from English-speaking jurisdictions, and is further fine-tuned with GDPR-specific datasets. This approach not only deepens LegiLM's mastery of legal terminologies and subtleties but also equips it to effectively tackle the specialized requirements of GDPR compliance assessments. Integrating cutting-edge NLP technologies with tailored legal insights, LegiLM signifies a notable advancement in legal AI technologies, aiming to streamline compliance processes and reduce the burden on legal practitioners.
Our contributions can be summarized as follows:

\begin{itemize}
    \item We developed LegiLM, a legal language model and specifically designed for Legal Question Answering (LQA) with a focus on GDPR compliance. This model is fine-tuned for the nuanced domain of legal compliance, leveraging a comprehensive dataset that includes GDPR regulations, annotated contracts, and related privacy policies. This allows LegiLM to accurately assess compliance issues with a high degree of legal reasoning and retrieval augmentation, particularly in accurately assessing compliance issues in data interaction-related behaviors, events, and contracts.

    \item The development of LegiLM is supported by the creation of a comprehensive dataset, including GDPR regulations, case law, and actual data-sharing contracts. This dataset, combined with a detailed fine-tuning process, forms a solid foundation for training the model. This method improves the model's precision in identifying compliance issues and maintains its relevance to contemporary legal standards and practices.

    \item Through rigorous benchmarking against other models, LegiLM has demonstrated effectiveness in interpreting complex legal documents and assessing GDPR compliance. This validation supports the efficacy of our domain-specific training approach and underscores LegiLM's utility in legal tech.
\end{itemize}

\section{Related Work}
The development of Chinese Legal Large Language Models (LLMs) within Natural Language Processing (NLP) primarily focused on legal domains, initially concentrated on crafting specialized models for specific tasks such as legal judgment prediction \cite{Ge2021,Huang2021,Cui2023b}. The intricate nature of legal systems and nuanced demands of legal reasoning have necessitated a shift toward comprehensive, domain-specific LLMs designed to address a broad spectrum of legal scenarios. Recent advancements led to the creation of models like LawGPT and DISC-LawLLM \cite{Yue2023,Zhou2024}, which engage in general legal consulting and complex tasks such as legal examination preparation and sophisticated legal reasoning, leveraging domain-specific knowledge through pre-training strategies with large-scale Chinese legal documents followed by supervised fine-tuning on legal datasets \cite{Zhou2024}. The integration of external knowledge bases in models like ChatLaw \cite{Cui2023b} and Legal-LM \cite{Shi2024b} enhances their reliability and accuracy, addressing data veracity and the hallucination problems with mechanisms like Mixture-of-Experts (MoE) and knowledge graphs to refine outputs for legal reasoning and consultation. The operationalization of these LLMs benefits from novel architectural features and learning methodologies, such as SAILER's structure-aware approach to legal case retrieval and LegalGPT's multi-agent framework which integrates function-specific legal chains of thought, enhancing model efficiency and facilitating dynamic engagement with legal tasks \cite{Li2023, Shi2024b}. This evolution not only demonstrates significant technical progression but also aligns with the logical frameworks and practical needs of the legal domain, heralding a new era of AI-enhanced legal services attuned to domain-specific complexities. 

In contrast, the English-speaking legal domain's LLMs, though less explored, have begun to carve a niche within NLP. Early contributions like LEGAL-BERT \cite{Chalkidis2020} adapted the BERT architecture to accommodate the unique terminologies and stylistic nuances of legal documents. Subsequent efforts like LexGLUE \cite{Chalkidis2022a} introduced a benchmarking dataset that unified various legal NLP tasks, facilitating comprehensive assessment across a spectrum of legal understanding challenges. The Cambridge Law Corpus \cite{Ostling2024} enriched resources with a dataset of UK court cases annotated for outcome predictions, supporting models capable of deep legal analysis. SaulLM-7B \cite{Colombo2024} epitomizes the progression toward highly specialized English legal LLMs with a training regime on an extensive legal corpus, underscoring capabilities in sophisticated legal text comprehension and generation, reflecting the growing acknowledgment of ethical dimensions necessary when deploying AI in sensitive legal contexts.

\section{Methodology}
\subsection{LegiLM}
LegiLM is a legal large language model specifically designed for English legal applications and contract analysis. It is a specialized legal language model derived from the foundational SaulLM-7B model, specifically tailored for sophisticated legal language processing and meticulous contract analysis. Engineered to enhance the automatic detection and interpretation of privacy and compliance clauses within data-sharing contracts, LegiLM leverages an enriched training regimen focused on data protection laws, privacy policies, and specific GDPR cases. The model is meticulously fine-tuned to assess compliance with current data protection regulations effectively. This fine-tuning process incorporates real contract instructions and GDPR provisions, significantly improving the model's capability to understand and process complex legal texts. Through this advanced adaptation, LegiLM supports businesses and legal professionals by providing precise insights and reliable guidance, ensuring that data processing activities align with legal standards.
\begin{figure}[H]
\centering
\includegraphics[width=0.9\textwidth]{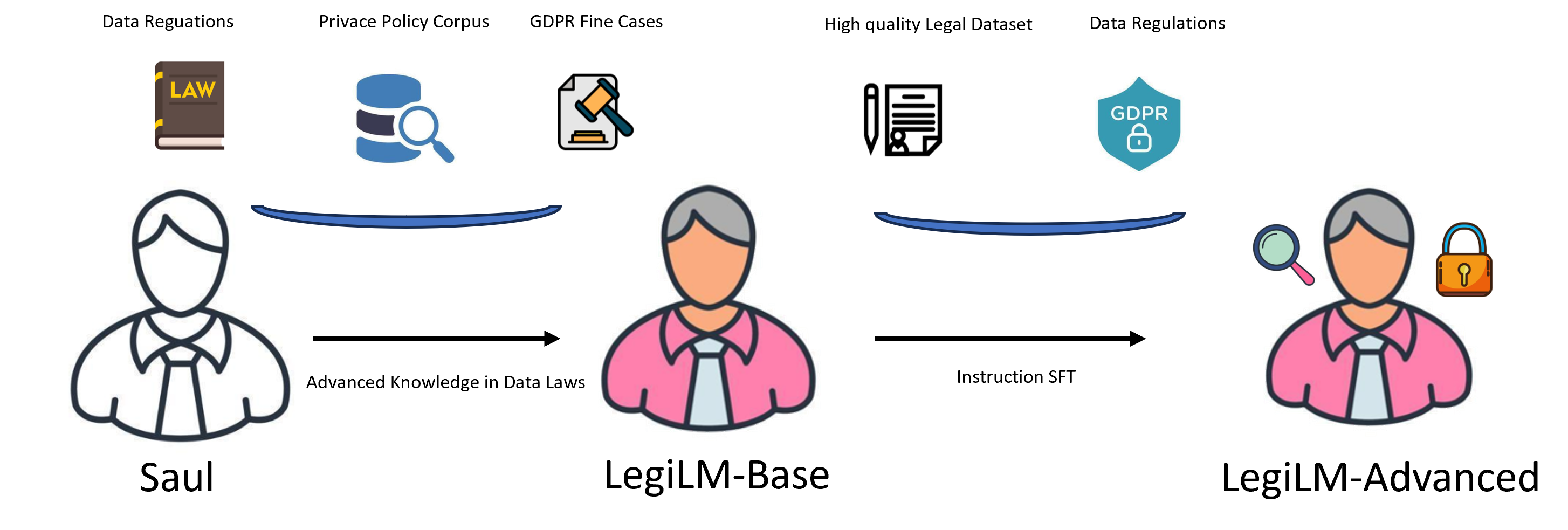}
\caption{ Procedure for constructing LegiLM.  }\label{fig:fig1}
\end{figure}
we have employed a comprehensive dataset that includes Data Regulations, the Privacy Policy Corpus, and cases pertaining to GDPR fines. The foundational training of LegiLM-Base was enhanced by incorporating deep legal insights from the data law sector, drawing on the specialized capabilities of SaulLM-7B, which is tailored for the legal domain. The initial phase of model training, designated as LegiLM-Base, was conducted using advanced insights derived from data law. This foundational training was crucial for the model to acquire a nuanced understanding of privacy policies and data regulations. Subsequently, we engaged in a more focused fine-tuning process utilizing instructions from Data Trading Contracts and specific GDPR legal provisions. This second phase of training, aimed at producing LegiLM-Advanced, was guided by a supervised fine-tuning methodology using data contract instructions. The rigorous training regimen ensured that LegiLM-Advanced emerged as a proficient tool in interpreting and complying with the complex legal frameworks associated with data regulations within the context of prompt.

\begin{tcolorbox}[colframe=darkgreen, colback=white, coltitle=white, title=\textbf{Question:Legal Consulting}]
\textbf{The agreement between DataCorp Ltd. and HealthAnalytics Inc. involves sharing anonymized health data for research purposes. It is effective from January 1, 2024, to December 31, 2025. Data will be transferred monthly using encrypted IPFS, accessible only to GDPR-trained personnel. The cost is 1000 USD per month, with specific invoicing and late payment terms. Is this agreement valid?} \\
\tcblower
\textit{Answer:} \textbf No\\
\textit{Output:} 
The Data Sharing Agreement meets GDPR requirements, notably through compliance with Article 32, which requires appropriate security measures. The agreement includes data anonymization, secure data transfers using IPFS with encryption, and access limited to GDPR-trained personnel. It also adheres to broader GDPR principles, such as lawful processing and data subject rights protection. Assuming consistent enforcement of these measures, the agreement is considered valid under GDPR, demonstrating the parties' commitment to data privacy and security, thereby affirming its legitimacy in adhering to GDPR standards.

\end{tcolorbox}
(a)Enhancing Legal Consulting Accuracy through Domain-Specific Language Model Application

The task of legal consulting frequently requires a deep understanding of domain-specific regulations and practices. General language models often lack this nuanced legal knowledge, making them less effective in addressing complex legal inquiries. By incorporating LegiLM, a fine-tuned legal language model for GDPR compliance detection in data-sharing contracts, we aim to bridge this gap. LegiLM leverages extensive legal domain knowledge and training on specific legal tasks to provide more accurate and contextually relevant advice, as demonstrated in the analysis of the DataCorp Ltd. and HealthAnalytics Inc. agreement.
\begin{tcolorbox}[colframe=blue, colback=white, coltitle=white, title=\textbf{Legal Clause Retriever}, sharp corners]
\textbf{The agreement between DataCorp Ltd. and HealthAnalytics Inc. involves sharing anonymized health data for research purposes. It is effective from January 1, 2024, to December 31, 2025. Data will be transferred monthly using encrypted IPFS, accessible only to GDPR-trained personnel. The cost is 10 USD per month, with specific invoicing and late payment terms. Is this agreement valid?} \\
\tcblower
\textit{Premise:} \\
Article 32 of the General Data Protection Regulation (GDPR), titled "Security of processing," lays out requirements for the security measures that must be taken when processing personal data. The relevant measures include:
\begin{itemize}
    \item[a.] The pseudonymisation and encryption of personal data;
    \item[b.] The ability to ensure the ongoing confidentiality, integrity, availability, and resilience of processing systems and services;
    \item[c.] The ability to restore the availability and access to personal data in a timely manner in the event of a physical or technical incident;
    \item[d.] A process for regularly testing, assessing, and evaluating the effectiveness of technical and organizational measures for ensuring the security of the processing.
\end{itemize}
\end{tcolorbox}
(b)Optimizing Legal Document Analysis with Targeted Clause Retrieval Capabilities

Legal document review processes can be lengthy and prone to human error, particularly when identifying specific compliance-related clauses. Traditional models may not effectively pinpoint relevant legal clauses due to insufficient specialized training. By employing LegiLM, our model specifically addresses these challenges by retrieving pertinent clauses, such as those highlighted in GDPR's Article 32 regarding security of processing. This model, refined through legal-oriented pre-training and supervised fine-tuning, excels at extracting and analyzing crucial legal information, thereby enhancing both efficiency and reliability in legal applications.

\subsection{Data Collection and Preparation}

\subsubsection{GDPR-Related Texts}

We compiled a comprehensive dataset from various authoritative sources, ensuring that it encompasses the full spectrum of GDPR-related legal content:

\begin{itemize}
\item \textbf{Data Regulations:} The full text of the GDPR\footnote{https://gdpr-info.eu/} was sourced from the official European Commission’s GDPR portal. Additionally, the text of the California Consumer Privacy Act (CCPA)\footnote{https://www.oag.ca.gov/privacy/ccpa} was obtained from the official website of the California Attorney General. This section includes all articles, recitals, and accompanying guidelines for both regulations.
\item \textbf{Legal Interpretations and Guidelines:} Additional documents, including guidelines from the European Data Protection Board (EDPB) and legal interpretations from EUR-Lex\footnote{https://eur-lex.europa.eu/ homepage.html}, were included to provide context and depth for specific GDPR articles.

\item \textbf{Case Law:} Relevant case law, including judgments from the Court of Justice of the European Union (CJEU) and national courts across the EU\footnote{https://eur-lex.europa.eu/browse/directories/new-case-law.html}, was collected to illustrate the application of GDPR in various legal scenarios.

\item \textbf{GDPR FINES Database:} A comprehensive database\footnote{https://www.privacyaffairs.com/gdpr-fines/} detailing fines imposed by data protection agencies across the EU for GDPR violations was incorporated to add context on enforcement actions and penalties.

\item \textbf{GDPR Enforcement Tracker Dataset:} GDPR Fines Tracker \footnote{https://www.enforcementtracker.com/} provides an overview of fines and penalties imposed by EU data protection agencies, offering insight into the practical enforcement of GDPR regulations.
\end{itemize}

\subsubsection{Data-Sharing Contracts and Privacy Policies}

\begin{itemize}
\item \textbf{Data-Sharing Contracts:} A comprehensive dataset of data-sharing legal contracts was developed using customizable templates tailored for GDPR compliance. Each contract has been meticulously crafted and annotated to reflect specific compliance or non-compliance with designated GDPR articles.
\item \textbf{Additional Privacy Policies:} To broaden the model's understanding of privacy regulations, additional privacy policies were included. These were sourced from company websites, legal advice platforms\footnote{https://policies.google.com/privacy,https://x.com/en/privacy,https://www.facebook.com/privacy/policy/}, and privacy policy repositories, contributing an additional 300,000 tokens.
\end{itemize}

\subsection{Data Preprocessing}

The preprocessing steps were designed to ensure the consistency and quality of the data:

\begin{itemize}
\item \textbf{Text Normalization:} All text data underwent normalization to remove artifacts such as HTML tags, special characters, and inconsistent formatting. Legal terminology was standardized to maintain uniformity across the dataset.
\item \textbf{Tokenization:} The dataset was tokenized using a legal-domain-specific tokenizer to preserve the integrity of legal phrases and ensure accurate parsing during model training.

\item \textbf{Annotation and Labeling:} Each contract and policy was thoroughly annotated with labels indicating GDPR compliance status. Specific references to GDPR articles were added to guide the model in linking legal provisions to contract clauses. Annotations also included specific violations like improper data processing, lack of data subject rights, and insufficient legal basis for data sharing.
\end{itemize}

\subsection{Legal Consultation Dataset Fine Tuning}
When fine-tuning language models to handle queries in the legal domain concerning digital information, we implemented a learning method based on a case-specific dataset. A comprehensive database detailing fines imposed by data protection agencies across the EU for data regulation violations was integrated to enrich the model with context on enforcement actions and penalties. This method encompasses the following essential steps:
\begin{itemize}
    \item \textbf{Instruction Construction:} Instructions related to specific legal issues are developed. The goal here is to equip the model with a detailed scenario that includes the legal actions under assessment and their potential regulatory conflicts. These instructions facilitate a thorough comprehension by the model of the context before it provides legal advice.
    \item \textbf{Answer Formalization:} The response to the legal queries in the instructions is typically straightforward, expressed as ``Yes'' or ``No''. This format helps in evaluating the model’s proficiency in understanding and applying legal knowledge accurately.
    \item \textbf{Output Elaboration:} This section offers an in-depth explanation of the responses, elucidating why certain behaviors are either in compliance with or in violation of the applicable laws. This training phase is vital, showcasing the model's ability to handle legal nuances and its competence in logical reasoning and legal interpretation.
\end{itemize}
\subsubsection{Maintaining Answer Diversity:} Maintaining a diversity of answers during the fine-tuning process is essential to avoid bias and overfitting toward certain response tendencies. Especially in the field of legal consultation, a pattern of providing the same type of answer, such as repeatedly saying ``Yes,'' can predispose the model to one category of response, overlooking the uniqueness and complexity of cases. We use Chatgpt to generating negative cases, even in datasets that predominantly contain positive examples, involves creating scenarios that clearly do not violate data regulations. This is accomplished through a contrastive learning approach, where each case is paired with one or more counterexamples. These counterexamples differ subtly on key legal points, leading to different compliance assessments, and help the model to learn finer discrimination standards, thus improving its sensitivity and judgment capabilities regarding legal details.
    \begin{tcolorbox}[colframe=darkgreen, colback=white, coltitle=white, title=\textbf{Original Template}]
\begin{verbatim}
{
  "instruction": "INGENIERIA Y TELECOM JAEN, S.L. extended a contract without 
  the consent of the data subject. Is this action compliant with GDPR?",
  "answer": "No",
  "output": "The action taken by INGENIERIA Y TELECOM JAEN, S.L. violates 
  GDPR Article 6, which mandates a lawful basis for any data processing. 
  Extending a contract without explicit consent from the data subject 
  bypasses the legal prerequisites for processing personal data."
}
\end{verbatim}\\
\end{tcolorbox}
 \begin{tcolorbox}[ colback=white, coltitle=white, title=\textbf{Contrastive Template}]
\begin{verbatim}
{
  "instruction": "INGENIERIA Y TELECOM JAEN, S.L. extended a contract after 
  automatically renewing the consent of the data subject via an opt-out 
  mechanism provided in the initial agreement. Is this action compliant 
  with GDPR?",
  "answer": "Yes",
  "output": "The action taken by INGENIERIA Y TELECOM JAEN, S.L. is compliant 
  with GDPR Article 6. The contract extension under the terms of an opt-out 
  consent renewal mechanism, which was clearly stipulated in the initial 
  agreement and to which the data subject did not object, provides a lawful 
  basis for processing personal data."
}
\end{verbatim}\\
\end{tcolorbox}
To construct a contrastive example, initiate by developing a scenario that closely mirrors the initial case but incorporates pivotal differences that alter the legal outcome from a "Yes" (violation) to a "No" (compliance). Following this, integrate both the initial and the contrastive examples into the training corpus. This method trains the model to discern subtle distinctions in GDPR compliance based on minimal variations within the scenarios. For instance, the model learns that not all extensions of contracts without fresh explicit consent are unlawful; crucially, the context of consent, such as the presence of opt-out mechanisms, plays a vital role. Furthermore, the model is refined to enhance its capability to comprehend and reason through the nuances of legal compliance across similar yet legally distinct situations. Subsequently, assess the model’s performance using a validation set that includes a balanced assortment of "Yes" and "No" outcomes across varied nuanced scenarios, a step essential to confirm that the model does not perpetuate any prior biases towards "Yes" answers.
\section{Evaluation}

\subsection{Benchmarking LegiLM}

To evaluate LegiLM’s effectiveness in GDPR compliance detection, we developed a custom benchmark dataset derived from the collected and annotated data:

\begin{itemize}
\item \textbf{Multiple-Choice Questions:} The dataset includes 200 multiple-choice questions extracted from annotated data-sharing contracts. These questions are designed to test the model's ability to identify specific GDPR compliance issues based on contract clauses.

\item \textbf{Open-Ended Questions:} We included 150 open-ended questions requiring the model to generate detailed explanations for whether a contract complies with GDPR, referencing specific articles and legal interpretations.

\item \textbf{Real-World Case Studies:} The benchmark features 50 complex data-sharing scenarios where the model must assess overall GDPR compliance and propose corrective actions. These case studies are intended to simulate real legal consultation tasks.
\end{itemize}

\subsection{Evaluation Metrics}

LegiLM’s performance on this benchmark was assessed using the following metrics:

\begin{itemize}
    \item \textbf{Compliance Justification Quality:} This qualitative metric, evaluated by legal experts, assesses the relevance and depth of the explanations provided by the model in open-ended questions and real-world case studies. It measures how well the model justifies its compliance decisions.
    
     \item \textbf{Legal Question Answering (LQA) Accuracy:} This metric measures the model's ability to correctly answer legal questions, particularly those related to data trading contracts and GDPR compliance. The accuracy is assessed by comparing the model’s responses against a set of pre-defined correct answers, including those derived from GDPR-related queries. This reflects the model's proficiency in understanding and processing legal texts within these specific domains and its capability to provide accurate, GDPR-compliant, and legally sound answers.
\end{itemize}

\subsection{Results}

LegiLM was benchmarked against other prominent legal and general-purpose language models, with the results summarized in Table 1.

\begin{table}[htbp]
\centering
\caption{Performance Metrics on LegalBench-Instruct for LegiLM and Baseline Models}
\begin{tabular}{lccc}
\hline
\textbf{Model} & \textbf{Accuracy} & \textbf{F1-Score} & \textbf{Justification Quality} \\
\hline
Qwen-1.5-72B  & 31.41\% & 33.87 & Medium \\
GPT-4  & 42.35\% & 43.85 & Medium \\
Lawyer-LLaMA-13B & 45.32\% & 23.02 & Medium \\
ChatLaw-13B  & 42.76\% & 42.63 & Medium \\
DISC-Law-7B  & 41.60\% & 43.99 & Medium \\
InternLM-Law-7B  & 46.71\% & 48.67 & Medium \\
Saul-7B  & 62.10\% & 63.15 & High \\
LegiLM-Advanced (Ours) & \textbf{68.05\%} & \textbf{68.21} & \textbf{High} \\
\hline
\end{tabular}
\end{table}

The evaluation results is for English data-sharing contracts, particularly in GDPR compliance tasks, with top performance in accuracy, F1-score, and justification quality. Saul-7B follows closely, excelling in general English legal tasks but slightly behind LegiLM-Advanced in this specialized area. While legal-specific models like InternLM-Law-7B perform well in their respective Chinese legal domains, they do not match the effectiveness of LegiLM-Advanced and Saul in English legal contexts. General-purpose models such as GPT-4 and Qwen-1.5-72B demonstrate moderate performance, underscoring the benefits of domain-specific fine-tuning that LegiLM-Advanced exemplifies.

\begin{figure}[H]
\centering
\includegraphics[width=0.9\textwidth]{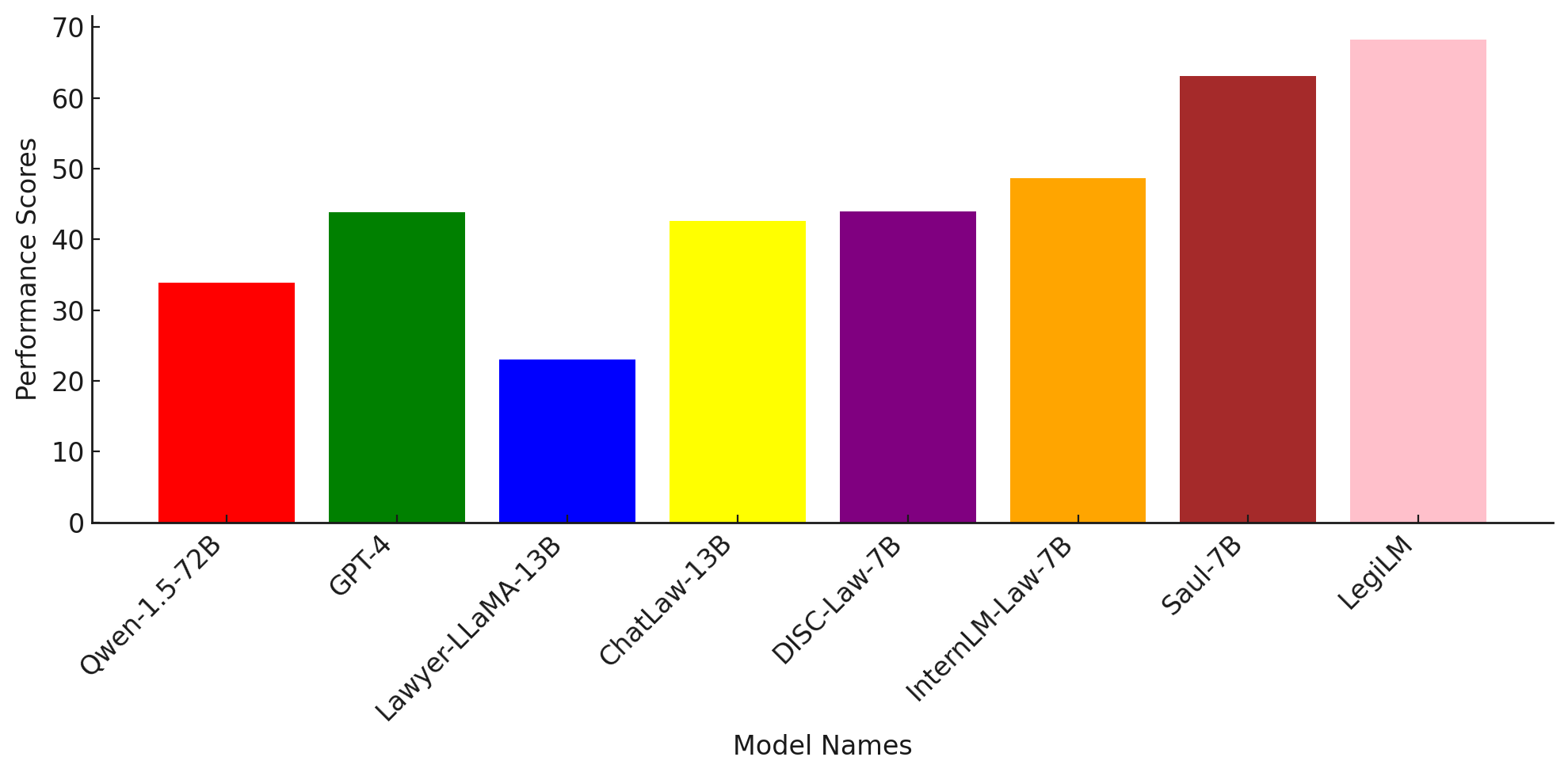}
\caption{ Performance Of Various Models In Legal Question Answering task}\label{fig:fig2}
\end{figure}

The performance in legal question answering task \ref{fig:fig2} demonstrates that domain-specific models, especially LegiLM and Saul-7B, significantly outperform others in the Legal Question Answering task, achieving scores of 68.21 and 63.15 respectively. LegiLM, a specialized legal language model derived from the SaulLM-7B model, is finely tuned for advanced legal language processing and contract analysis, particularly in the context of data protection laws and GDPR compliance. This fine-tuning, which incorporates real contract instructions and GDPR provisions, enables LegiLM to excel in detecting and interpreting privacy and compliance clauses within data-sharing contracts. In contrast, general-purpose models like GPT-4 perform moderately well, while less specialized models. Models like Lawyer-LLaMA-13B and Qwen-1.5-72B, which are Chinese language models, lag behind. This underscores the critical importance of both domain-specific fine-tuning and language specialization to achieve superior accuracy in specialized legal tasks.

\subsection{Discussion}

LegiLM’s performance, especially in the context of data compliance detection within privacy and data security, underscores the importance of domain-specific fine-tuning. By leveraging a targeted dataset and incorporating sophisticated legal reasoning through instruction fine-tuning, LegiLM effectively bridges the gap between general legal language processing and specific legal tasks. This makes it a valuable tool for legal practitioners seeking to automate and streamline data protection regulation compliance assessments.

At present, LegiLM has been fine-tuned specifically for GDPR compliance. However, future work will focus on expanding its capabilities to cover data protection regulations across various countries and regions, enabling broader applicability in global legal contexts.
\section{Conclusion}
This study presents LegiLM, a legal language model designed for compliance detection in data interaction activities. By incorporating advanced NLP technologies and deep legal knowledge, LegiLM achieves better performance compared to existing models, marking a significant improvement in the field of legal AI. Its capability for thorough compliance analysis and legal justification aligns well with the rigorous standards of modern data protection regulations. Our research offers valuable contributions to legal technology by providing a tool that improves the accuracy and efficiency of compliance assessments. Implementing LegiLM could enhance legal practices by reducing the need for extensive manual reviews. Furthermore, the approaches and insights from this study may inform future AI advancements in the legal sector, with potential applicability across various regulatory frameworks and jurisdictions, enhancing AI's role in global legal systems.

\bibliographystyle{plainnat}  
\bibliography{LegalAI.bib}

\begin{thebibliography}{20}
\providecommand{\natexlab}[1]{#1}
\providecommand{\url}[1]{\texttt{#1}}
\expandafter\ifx\csname urlstyle\endcsname\relax
  \providecommand{\doi}[1]{doi: #1}\else
  \providecommand{\doi}{doi: \begingroup \urlstyle{rm}\Url}\fi

\bibitem[Chalkidis et~al.(2020)Chalkidis, Fergadiotis, Malakasiotis, Aletras, and Androutsopoulos]{Chalkidis2020}
Ilias Chalkidis, Manos Fergadiotis, Prodromos Malakasiotis, Nikolaos Aletras, and Ion Androutsopoulos.
\newblock Legal-bert: The muppets straight out of law school.
\newblock \emph{Findings of the Association for Computational Linguistics: EMNLP 2020}, pages 2898--2904, 2020.

\bibitem[Chalkidis et~al.(2022)Chalkidis, Jana, Hartung, Bommarito, Androutsopoulos, Katz, and Aletras]{Chalkidis2022a}
Ilias Chalkidis, Abhik Jana, Dirk Hartung, Michael Bommarito, Ion Androutsopoulos, Daniel Katz, and Nikolaos Aletras.
\newblock Lexglue: A benchmark dataset for legal language understanding in english.
\newblock In \emph{Proceedings of the 60th Annual Meeting of the Association for Computational Linguistics}, pages 4310--4330, 2022.

\bibitem[Chen and Zhang(2013)]{Chen2013}
Wei Chen and Xiaoyang Zhang.
\newblock Legal document retrieval: A benchmark study.
\newblock \emph{Information Retrieval Journal}, 16:\penalty0 145--162, 2013.

\bibitem[Colombo et~al.(2024)Colombo, Pires, Boudiaf, Culver, Melo, Corro, and Desa]{Colombo2024}
P.~Colombo, T.~P. Pires, M.~Boudiaf, D.~Culver, R.~Melo, C.~Corro, and M.~Desa.
\newblock Saullm-7b: A pioneering large language model for law.
\newblock \emph{arXiv preprint arXiv:2403.03883}, 2024.

\bibitem[Cui et~al.(2023)Cui, Li, Yan, Chen, and Yuan]{Cui2023b}
Jiaxi Cui, Zongjian Li, Yang Yan, Bohua Chen, and Li~Yuan.
\newblock Chatlaw: Open-source legal large language model with integrated external knowledge bases.
\newblock \emph{arXiv preprint arXiv:2306.16092}, 2023.

\bibitem[Du et~al.(2022)Du, Qian, Liu, Ding, Qiu, Yang, and Tang]{Du2022}
Zhengxiao Du, Yujie Qian, Xiao Liu, Ming Ding, Jiezhong Qiu, Zhilin Yang, and Jie Tang.
\newblock Glm: General language model pretraining with autoregressive blank infilling.
\newblock In \emph{Proceedings of the 60th Annual Meeting of the Association for Computational Linguistics (Volume 1: Long Papers)}, pages 320--335, 2022.

\bibitem[Ge et~al.(2021)Ge, Huang, Shen, Li, and Hu]{Ge2021}
Jidong Ge, Yunyun Huang, Xiaoyu Shen, Chuanyi Li, and Wei Hu.
\newblock Learning fine-grained fact-article correspondence in legal cases.
\newblock \emph{IEEE/ACM Transactions on Audio, Speech, and Language Processing}, 29:\penalty0 3694--3706, 2021.

\bibitem[Huang et~al.(2021)Huang, Shen, Li, Ge, and Luo]{Huang2021}
Yunyun Huang, Xiaoyu Shen, Chuanyi Li, Jidong Ge, and Bin Luo.
\newblock Dependency learning for legal judgment prediction with a unified text-to-text transformer.
\newblock \emph{arXiv preprint arXiv:2112.06370}, 2021.

\bibitem[Li et~al.(2023)Li, Ai, Chen, Dong, Wu, Liu, and Tian]{Li2023}
H.~Li, Q.~Ai, J.~Chen, Q.~Dong, Y.~Wu, Y.~Liu, and Q.~Tian.
\newblock Sailer: Structure-aware pre-trained language model for legal case retrieval.
\newblock In \emph{Proceedings of the 46th International ACM SIGIR Conference on Research and Development in Information Retrieval}, pages 1035--1044, 2023.

\bibitem[Licari and Comandè(2022)]{Licari2022}
John Licari and Giuseppe Comandè.
\newblock Legal language models: From general to specialized applications.
\newblock \emph{Artificial Intelligence Review}, 55:\penalty0 467--482, 2022.

\bibitem[Martin and Carter(2024)]{Martin2024}
Emily Martin and Andrew Carter.
\newblock Challenges and opportunities in legal ai: A comprehensive survey.
\newblock \emph{Journal of Legal Technology}, 12:\penalty0 220--238, 2024.

\bibitem[Muennighoff et~al.(2023)Muennighoff, Zhang, and Chen]{Muennighoff2023}
Nils Muennighoff, Wei Zhang, and Rui Chen.
\newblock Efficient code generation using pretrained language models.
\newblock \emph{ACM Transactions on Software Engineering and Methodology}, 32:\penalty0 78--90, 2023.

\bibitem[OpenAI(2023)]{OpenAI2023}
OpenAI.
\newblock Gpt-4 technical report.
\newblock \emph{arXiv preprint}, 2303:\penalty0 08774, 2023.

\bibitem[{\"O}stling et~al.(2024){\"O}stling, Sargeant, Xie, Bull, Terenin, Jonsson, and Steffek]{Ostling2024}
A.~{\"O}stling, H.~Sargeant, H.~Xie, L.~Bull, A.~Terenin, L.~Jonsson, and F.~Steffek.
\newblock The cambridge law corpus: a dataset for legal ai research.
\newblock Technical Report~11, University of Cambridge Faculty of Law Research Paper, 2024.

\bibitem[Shi et~al.(2024)Shi, Guo, Liao, Wang, Chen, and Liang]{Shi2024b}
J.~Shi, Q.~Guo, Y.~Liao, Y.~Wang, S.~Chen, and S.~Liang.
\newblock Legal-lm: Knowledge graph enhanced large language models for law consulting.
\newblock In \emph{International Conference on Intelligent Computing}, 2024.

\bibitem[Singhal et~al.(2023)Singhal, Kumar, and Singh]{Singhal2023}
Ankur Singhal, Raj Kumar, and Neelam Singh.
\newblock Large language models for healthcare: A review.
\newblock \emph{Journal of Healthcare Informatics}, 15:\penalty0 45--60, 2023.

\bibitem[Tang et~al.(2021)Tang, Liu, and Chen]{Tang2021}
Hongyu Tang, Jie Liu, and Xia Chen.
\newblock Advancements in coding with large language models.
\newblock \emph{Proceedings of the IEEE Conference on Computer Vision and Pattern Recognition}, pages 1234--1241, 2021.

\bibitem[Touvron et~al.(2023)Touvron, Lavril, Izacard, Martinet, Lachaux, Lacroix, Rozière, Goyal, Hambro, Azhar, Rodriguez, Joulin, Grave, and Lample]{Touvron2023}
Hugo Touvron, Thibaut Lavril, Gautier Izacard, Xavier Martinet, Marie-Anne Lachaux, Timothée Lacroix, Baptiste Rozière, Naman Goyal, Eric Hambro, Faisal Azhar, Aurélien Rodriguez, Armand Joulin, Edouard Grave, and Guillaume Lample.
\newblock Llama: Open and efficient foundation language models.
\newblock \emph{CoRR}, abs/2302.13971, 2023.

\bibitem[Yue et~al.(2023)Yue, Li, and Zhang]{Yue2023}
Zhenyu Yue, Mingyu Li, and Jie Zhang.
\newblock Lawgpt: A legal language model for chinese legal intelligence.
\newblock \emph{arXiv preprint}, 2023.

\bibitem[Zhou et~al.(2024)Zhou, Zhang, and Liu]{Zhou2024}
Yifan Zhou, Yifan Zhang, and Wei Liu.
\newblock Disc-lawllm: A domain-specific large language model for legal reasoning.
\newblock \emph{Journal of Artificial Intelligence Research}, 77:\penalty0 120--135, 2024.

\end{thebibliography}

\end{document}